\documentclass[letterpaper,10pt,conference]{IEEEtran}

\IEEEoverridecommandlockouts

\usepackage{cite}
\usepackage{amsmath,amssymb,amsfonts}
\usepackage{algorithmic}
\usepackage{graphicx}
\usepackage{textcomp}
\usepackage{verbatim}
\usepackage{booktabs}
\usepackage{xcolor}

\def\BibTeX{{\rm B\kern-.05em{\sc i\kern-.025em b}\kern-.08em
		T\kern-.1667em\lower.7ex\hbox{E}\kern-.125emX}}

\begin{document}
\setlength{\columnsep}{0.25in}
\title{LLM-Empowered Multimodal Fusion Framework for Autonomous Driving: Semantic Enhancement and Channel-Adaptive Design}

\author{
	\IEEEauthorblockN{Wen Wang$^{1,2}$,Yaping Sun$^{1,3}$,Yejun He$^{2}$,
Hao Chen$^{1}$,Zhiyong Chen$^{4}$,Xiaodong Xu$^{5,1}$,Nan Ma$^{5,1}$,Shuguang Cui$^{3}$}
	\IEEEauthorblockA{$^{1}$ Department of Broadband Communication, Pengcheng Laboratory, Shenzhen, China}
	\IEEEauthorblockA{$^{2}$ College of Electronic and Information Engineering, Shenzhen University, Shenzhen, China}
	\IEEEauthorblockA{$^{3}$ Future Network of Intelligence Institute (FNii), the Chinese University of Hong Kong (Shenzhen), Shenzhen, China}
	\IEEEauthorblockA{$^{4}$  Cooperative Medianet Innovation Center, Shanghai Jiao Tong University, Shangha,China}
	\IEEEauthorblockA{$^{5}$ Beijing University of Posts and Telecommunications, Beijing, China}
	
	\IEEEauthorblockA{ wangwenstu@outlook.com, sunyp@pcl.ac.cn, heyejun@126.com, chenh03@pcl.ac.cn, \\
		zhiyongchen@sjtu.edu.cn, xuxd@pcl.ac.cn, manan@bupt.edu.cn, shuguangcui@cuhk.edu.cn }
	\thanks{This work was supported by the National Natural Science Foundation of China under Grant Nos. 62301471 and U2541208, the National Key Research and Development Program of China under Grant 2023YFE0107900, the Key Program of Shenzhen Natural Science Foundation under Grant JCYJ20241202124219023, and the Program of Shenzhen Key Laboratory Evaluation under Grant SYSPG20241211173908022.}
}
\maketitle

\begin{abstract}
	Vision--radar fusion is central to robust autonomous driving, combining dense visual semantics with precise range and velocity measurements from radar. However, real-world fusion quality is fundamentally challenged by dynamically varying input quality, stemming from occlusion, adverse weather, and channel noise. To address this, we re-frame the problem from static data fusion to channel-aware semantic reasoning and propose a Large Language Model-centric Semantic-layer Channel-aware Integrated Perception (LM-SCIP) framework. It places a Large Language Model (LLM) as a central reasoning core to fuse a local visual stream with a quality-varying external radar stream used to cover perception-blind spots. Concretely, LM-SCIP couples a hierarchical radar--vision encoder with a Channel-Adaptive Semantic Module (CASM) that maps link indicators into a ``Channel Prompt'' to dynamically gate external radar features. A parameter-efficient, LoRA-tuned LLM, in conjunction with a heterogeneous Mixture-of-Experts (H-MoE), then arbitrates between local visual cues and the channel-conditioned radar context. Finally, a decoupled multi-task decoder outputs localization, trajectory forecasting, and image reconstruction. Experiments on nuScenes and VIRAT validate our approach. On nuScenes, under a controlled toggle of radar input, LM-SCIP reduces localization RMSE by 40.0\% versus a vision-only baseline. On VIRAT, the model attains a 0.214m localization RMSE and 0.179m minFDE (k=1). These results reveal that the proposed LM-SCIP enables a robust vision-dominant fallback at low SNR and synergistic fusion at high SNR.
\end{abstract}


\section{Introduction}

\IEEEPARstart{M}{ultimodal} perception that fuses cameras and radar---often via Bird’s-Eye View (BEV) representations~\cite{ref1}---has become a cornerstone of robust autonomous driving. These systems leverage the complementary strengths of the two modalities: cameras offer rich semantics and texture, whereas radar provides precise range, velocity measurements and robustness in adverse weather~\cite{ref2}. The prevailing paradigm is deep, feature-level fusion---in which features from separate backbones are merged mid-stream---and BEV-transformer families have demonstrated state-of-the-art accuracy~\cite{ref3,ref4,ref5}.

\noindent\textbf{From single-vehicle to cooperative perception.} A fundamental limitation of single-vehicle systems is their susceptibility to occlusions. In complex urban scenes, static structures like buildings create blind regions that onboard sensors cannot fully observe. Cooperative perception has emerged as a powerful solution, extending the effective field of view by sharing sensing evidence among agents via V2X links (e.g., vehicle-to-infrastructure, V2I; vehicle-to-vehicle, V2V). However, this paradigm introduces a critical challenge beyond classical fusion: input quality becomes time-varying. External data streams conveyed over wireless links can degrade due to fluctuations in signal-to-noise ratio (SNR), interference, or packet loss. Most contemporary fusion stacks are fundamentally context-agnostic; designed under the assumption of reliable inputs, they employ fixed fusion policies and thus exhibit brittle behavior under real-world V2X dynamics~\cite{ref6}.

\begin{figure}[t]
	\centering
	\includegraphics[width=0.7\linewidth]{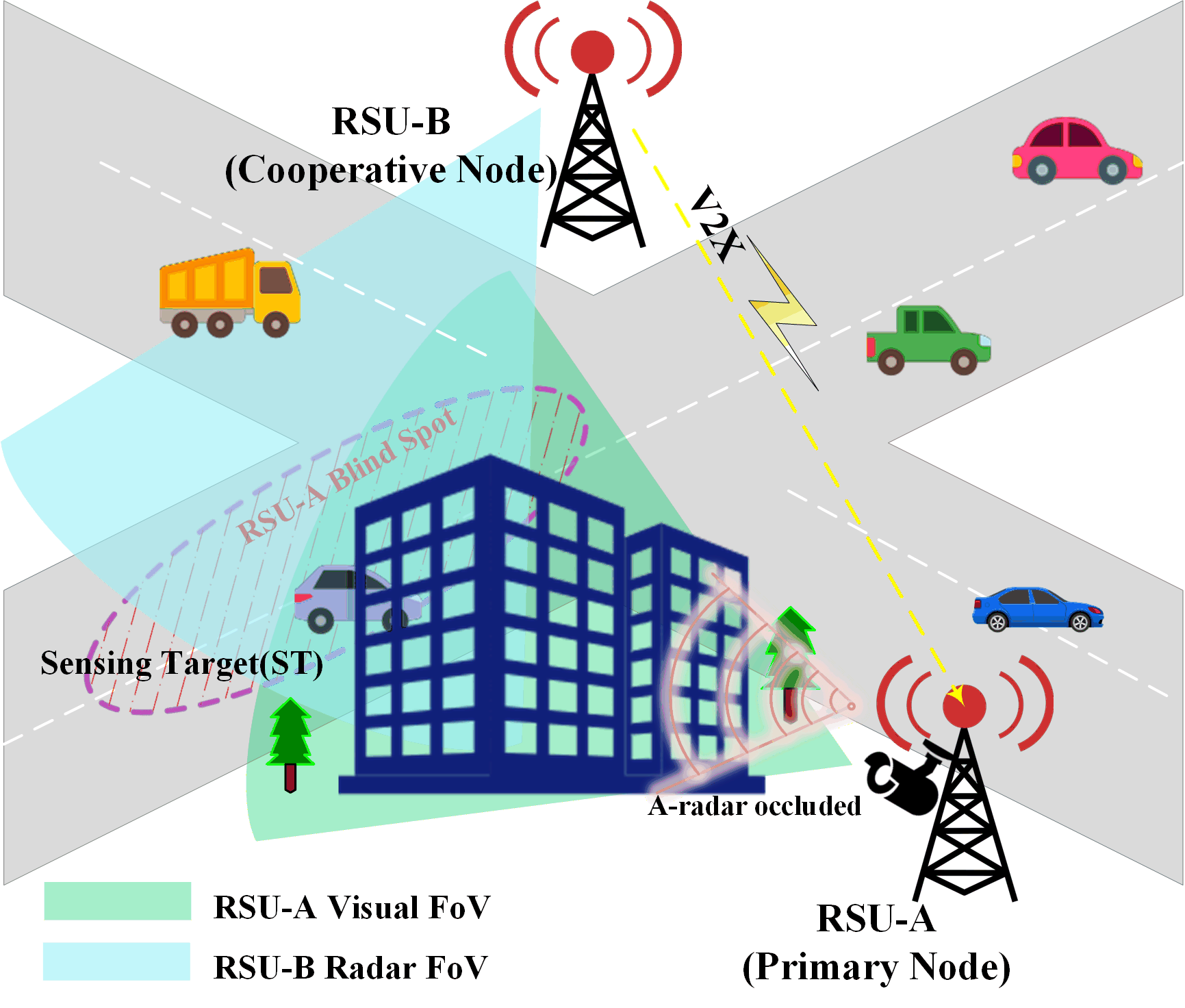} 
	\caption{Illustration of the cooperative perception scenario to resolve occlusion.}
	\label{fig1}
\end{figure}

\noindent\textbf{Our view.} We elevate fusion from the feature level to the \emph{semantic layer} and cast cooperative perception as \emph{channel-aware semantic reasoning}. We propose \emph{LM-SCIP}, an LLM-centric framework. \emph{LM-SCIP} injects instantaneous link quality into the perception loop: a CASM maps physical link indicators (e.g., SNR, modulation index) into a compact \emph{Channel Prompt} that conditions the in-loop LLM. The LLM then adapts the reliance on external radar versus local vision, producing a unified semantic representation for downstream multi-task perception.

\noindent\textbf{Positioning and distinctions.} Our work sits at the intersection of advanced BEV-style fusion and the burgeoning use of LLMs in autonomy. Traditional radar--vision fusion systems~\cite{ref3,ref4} and subsequent refinements~\cite{ref5} achieve strong accuracy under stable inputs, but typically assume high-quality sensor streams and implement fixed fusion policies, lacking explicit mechanisms to adapt to instantaneous quality fluctuations arising in cooperative settings. Recent efforts that incorporate LLMs into autonomy primarily treat them as decoupled, high-level planners operating on pre-processed or textual inputs~\cite{ref7,ref8,ref9}, which introduces an information bottleneck; end-to-end VLMs (e.g., DriveVLM~\cite{ref10}) mitigate discretization but remain unimodal. Unified perception--prediction--planning frameworks such as UniAD~\cite{ref11} improve holism, yet they generally do not couple LLM-style reasoning to link quality, and thus lack explicit \emph{link-aware} adaptation in multimodal fusion~\cite{ref12}. In short, prior art does not directly address the core challenge of cooperative perception: dynamically allocating trust across modalities when externally sourced inputs vary with wireless channel conditions.By contrast, \emph{LM-SCIP} treats link quality as a first-class context for perception and performs \emph{semantic-layer}, link-aware reasoning with an in-loop LLM, enabling real-time trust allocation across modalities under time-varying channels.

A parallel line of work in semantic communication for integrated sensing and communication (ISAC), exemplified by SIMAC~\cite{ref13}, optimizes over-the-air semantic coding and task-oriented decoders. Our focus is complementary: rather than designing a joint source--channel codec, \emph{LM-SCIP} targets robust, RSU-centric perception conditioned on the quality of an incoming cooperative link, using link indicators as side information to inform fusion and maintain reliability under time-varying channels.

\noindent\textbf{Contributions}---Our main contributions are:
\begin{itemize}
	\item[(i)] A channel-aware, LLM-centric paradigm for cooperative multimodal perception that performs \emph{semantic-layer} reasoning beyond static feature fusion.
	\item[(ii)] CASM for link-conditioned, feature-wise gating via a Channel Prompt, enabling robust performance under varying input quality.
	\item[(iii)] A LoRA-tuned LLM with a heterogeneous mixture of experts (H-MoE) and a decoupled decoder for stable, sample-efficient multi-task learning across localization, forecasting, and reconstruction.
\end{itemize}

\noindent\textbf{Empirical glimpse.} On nuScenes~\cite{ref14} and VIRAT~\cite{ref15}, our approach demonstrates strong fusion gains (e.g., large reductions in localization error versus vision-only), while ablations confirm that removing CASM or H-MoE markedly degrades performance under low sensing quality.

\section{System Model and Problem Formulation}\label{sec:system}

As shown in Fig.~\ref{fig1}, we consider an infrastructure-centric cooperative perception scenario centered on a primary roadside unit (RSU-A). RSU-A carries a co-located high-resolution camera and a radar. In an urban-corner layout, a building blocks the line-of-sight of RSU-A's radar toward part of the intersection, creating a \emph{radar blind region}. The local camera still provides a rich view but only partially observes targets behind the occluder.

\begin{figure*}[htbp]
	\centering
	\includegraphics[width=0.60\textwidth]{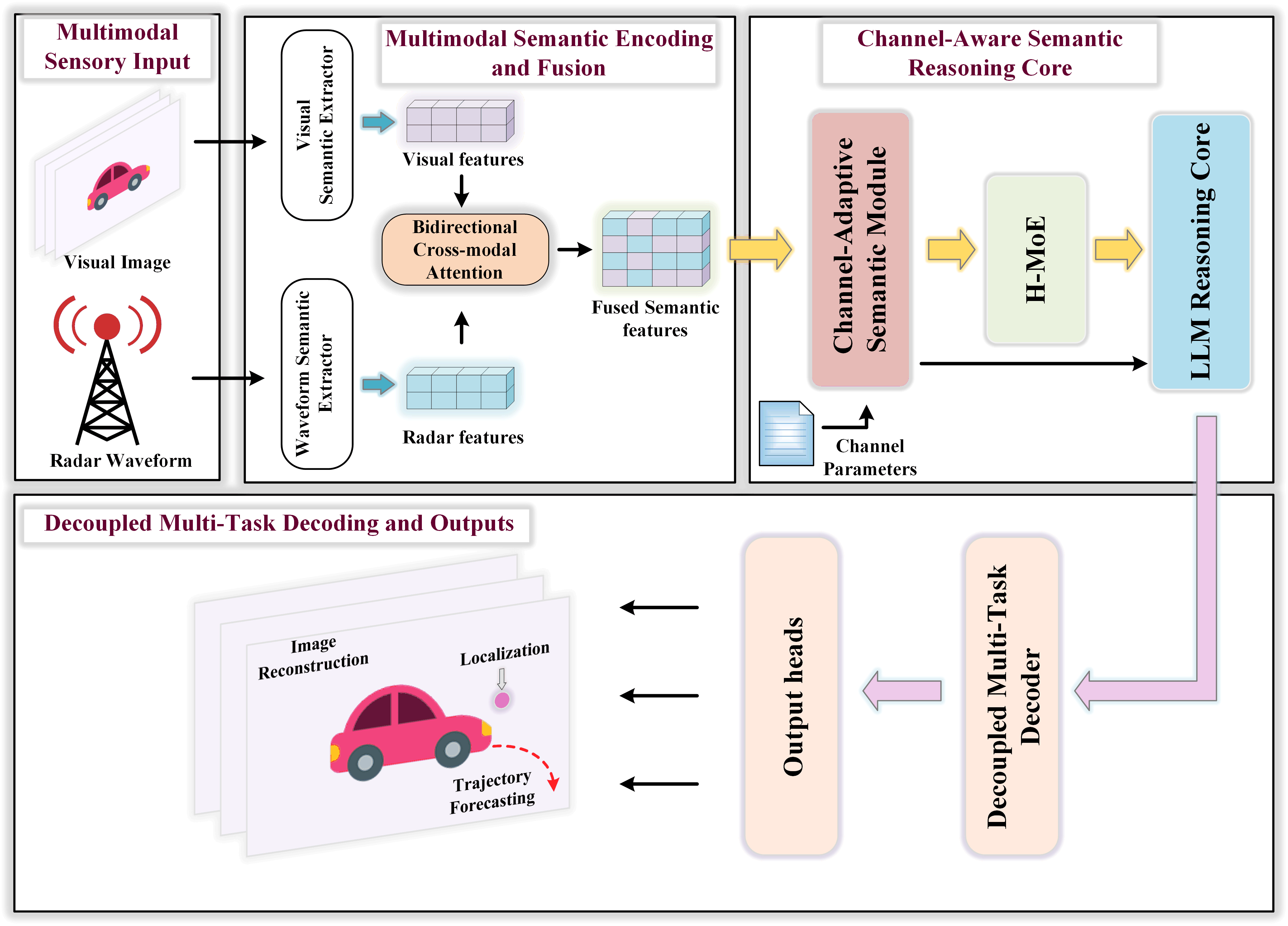}
	\caption{System overview of \emph{LM-SCIP}.}
	\label{fig2}
\end{figure*}

To close this gap, RSU-A leverages a cooperative node (another RSU or a connected vehicle; denoted RSU-B) whose field of view complements RSU-A and \emph{covers} the blind region with its radar. The proposed \emph{LM\mbox{-}SCIP} runs entirely on RSU-A's edge compute and fuses (i) \emph{local} visual stream  with (ii) a \emph{quality-varying external} radar stream that RSU-B \emph{transmits} to RSU-A over an RSU-to-RSU V2X sidelink~\cite{ref16}. We assume spatial and temporal calibration across nodes: all streams are time-stamped under a shared clock and mapped, via known extrinsics, into a common reference frame (RSU-A's camera frame) prior to fusion. For a given target $\mathrm{ST}_n$, the system estimates its location, future trajectory, and a high-fidelity visual reconstruction,as shown in Fig.~\ref{fig2}.

\subsection{System Setup and Sensing/Channel Modeling}
\noindent\textbf{Model inputs (LM\mbox{-}SCIP @ RSU-A).}
The onboard \emph{LM\mbox{-}SCIP} model consumes three inputs:
\begin{itemize}
	\item \textbf{Local visual stream.} The primary visual input is the image patch $m_n \in \mathbb{R}^{C \times H \times W}$ captured by RSU-A's co-located camera. We model this stream as clean and high-fidelity; residual imperfections stem from the physical environment (e.g., adverse weather). 
	
	\item \textbf{External radar stream (blind-spot coverage).} Because RSU-A's own radar is ineffective in the occluded sector, the cooperative node (RSU-B) first performs local radar sensing: it transmits an LFM waveform, receives the echo from a target $\mathrm{ST}_n$ within RSU-A's blind spot, then transmits the resulting raw complex I/Q waveform $\mathbf{A}^{(B)}_n(t)$ to RSU-A. The RSU-B to RSU-A transport is modeled as an \emph{equivalent} complex AWGN channel; thus RSU-A receives the \emph{noisy} waveform $\tilde{\mathbf{A}}_n(t)$ as as the input to the waveform encoder.
	
	\item \textbf{Channel-state indicators.} During reception of the external radar stream, RSU-A's modem reports instantaneous link indicators $\mathcal{J}_n=\{\mathrm{SNR},\,\mathrm{mod\_idx}_n\}$ , which are fed to CASM as side information.
\end{itemize}

\noindent\textbf{LFM--SIMO echo synthesis.}
Following the standard LFM--SIMO model~\cite{ref17}, the complex baseband echo received at the $k$-th antenna can be written as:
\begin{equation}
	x_{n,k}(t) \;=\; \lambda(d_n)\, a_k(\theta_n)\, s\!\bigl(t-\tau_n\bigr)\,
	\mathrm{e}^{\mathrm{j}2\pi \mu_n \bigl(t-\tau_n\bigr)},                                 
\end{equation}
stacking the $K$ channels gives the array echo:
\begin{equation}
	\mathbf{A}_n^{(B)}(t) \;=\; \lambda(d_n)\, \mathbf{a}(\theta_n)\, s\!\bigl(t-\tau_n\bigr)\,
	\mathrm{e}^{\mathrm{j}2\pi \mu_n \bigl(t-\tau_n\bigr)}.
	\label{eq:array_echo}
\end{equation}
where,
\begin{equation}
	\lambda(d_n) \;=\; \frac{\xi \cdot \rho_n}{(4\pi)^{3/2}\, d_n^{2}}, \qquad                     
	\xi \;=\; \frac{c}{f_c + K_t}.                                                            
\end{equation}
while $\rho_n$ is the radar cross section (RCS) of the $n$-th target, 
$\mathbf{a}(\theta_n)=\!\bigl[1,\,\mathrm{e}^{-\mathrm{j}\pi\cos\theta_n},\,\ldots,\,\mathrm{e}^{-\mathrm{j}(K-1)\pi\cos\theta_n}\bigr]^{\!\top}$ 
is the $K$-element ULA steering vector, $s(t)$ is the transmitted LFM baseband waveform,
$\tau_n = \tfrac{2 d_n}{c}$ is the round-trip delay for range $d_n$, and 
$\mu_n = \tfrac{2 v_n (f_c + K_t/2)}{c}$ is the Doppler frequency induced by radial velocity $v_n$.
Here $f_c$ is the carrier frequency, $K_t$ is the chirp rate, and $c$ is the speed of light.
By synthesizing $\mathbf{A}_n^{(B)}(t)$ from these physical parameters, we construct training inputs that enable the
\emph{Waveform Semantic Encoder} to learn the inverse mapping from raw waveform to a compact semantic state.

\noindent\textbf{Waveform-level channel (training emulation).}
To emulate residual transport distortion under latency-constrained delivery, we inject complex AWGN onto the synthesized echo and form the noisy input used by the encoder:
\begin{equation}
	\tilde{\mathbf{A}}_n(t) \;=\; \mathbf{A}_n^{(B)}(t) \;+\; \mathbf{N}(t),
	\label{eq:waveform}
\end{equation}
where the noise power $\sigma_N^2$ is set relative to the signal power $\sigma_A^2$ such that $\mathrm{SNR}=10\log_{10}(\sigma_A^2/\sigma_N^2)$. 

\noindent\textbf{Semantic-layer channel conditioning.}
Beyond waveform noise, CASM injects channel awareness at the semantic layer: the indicators $\mathcal{J}_n$ are embedded into a multi-token \emph{Channel Prompt} that dynamically gates the contribution of external-radar features before decoding, enabling vision-dominant fallback when link quality is poor and synergistic gains when the cooperative radar is reliable.
\subsection{Problem Formulation}
Let $\mathcal{D} = \{(\mathbf{m}_n, \tilde{\mathbf{A}}_n, \mathbf{y}_n)\}_{n=1}^{N}$ be the dataset, 
where 
$\mathbf{y}_n = \{\mathbf{P}_{\mathrm{loc}},\, \mathbf{P}_{\mathrm{traj}},\, \mathbf{m}_n\}$ 
collects the ground-truth labels for localisation, trajectory, and reconstruction. 
The end-to-end model is:
\begin{equation}
	\hat{\mathbf{y}}_n = F_{\Theta}\bigl(\mathbf{m}_n,\, \tilde{\mathbf{A}}_n,\, \mathcal{J}_n\bigr),
\end{equation}
where 
$\hat{\mathbf{y}}_n = \{\hat{\mathbf{P}}_{\mathrm{loc}},\, \hat{\mathbf{P}}_{\mathrm{traj},n},\, \hat{\mathbf{m}}_n\}$,
and $\Theta$ collects the parameters of the multimodal encoder, CASM, the LLM–H-MoE reasoning module, and the decoders. Training minimizes a multi-task objective over $\mathcal{D}$ and channel draws $\mathcal{J}_n \sim p(\mathcal{J})$:
\begin{equation}
	\Theta^{*} \;=\; \arg\min_{\Theta} \;
	\mathbb{E}_{(\mathbf{m}_n, \mathbf{A}_n, \mathbf{y}_n)\sim\mathcal{D}, \;\mathcal{J}_n \sim p(\mathcal{J})}
	\Bigl[\, \mathcal{L}_{\mathrm{total}}\bigl(\hat{\mathbf{y}}_n, \mathbf{y}_n\bigr) \,\Bigr],
	\label{eq:objective}
\end{equation}
where $\mathcal{L}_{\mathrm{total}}$ is a weighted sum of reconstruction, localization, trajectory, and auxiliary losses.

\section{The Proposed LM-SCIP Framework}

\subsection{Overview}
\begin{figure*}[!t]
	\centering
	\includegraphics[width=0.66\textwidth]{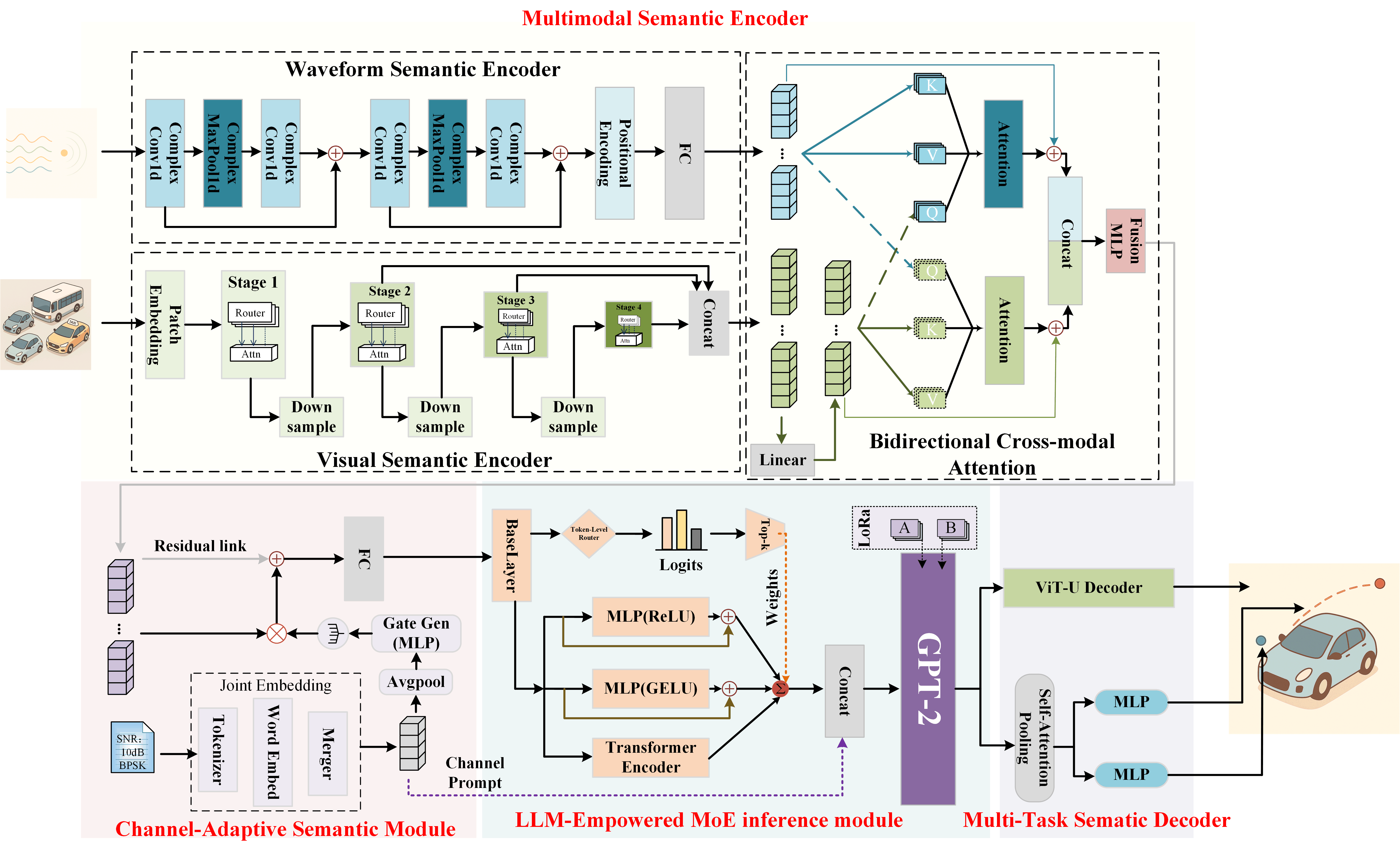}
	\caption{Network design of the proposed \emph{LM-SCIP} framework.}
	\label{fig3}
\end{figure*}

We elevate multimodal fusion to the \emph{semantic} layer and cast cooperative perception as \emph{channel-aware semantic reasoning}. As shown in Fig.~\ref{fig3}, \emph{LM\mbox{-}SCIP} comprises four stages: (i) a Multimodal Semantic Encoder (MSE) that forms a unified, time-aligned radar–vision token sequence; (ii) a Channel-Adaptive Semantic Module (CASM) that embeds V2X link indicators (SNR and modulation) into a compact \emph{Channel Prompt} to dynamically gate \emph{external-radar} features; (iii) an in-loop heterogeneous Mixture-of-Experts (H-MoE) with a LoRA-tuned LLM as the central reasoning core; and (iv) a decoupled multi-task decoder for image reconstruction (Recon), localization (Loc), and trajectory (Traj) forecasting.

\subsection{Multimodal Semantic Encoder (MSE)}
The MSE ingests an image patch and a complex radar waveform with modality-specific backbones (hierarchical ViT for vision; complex CNN for I/Q) and aligns them by a lightweight bidirectional cross-modal attention, yielding a compact sequence $\mathbf{S}^{\mathrm{multi}}$ for downstream reasoning. Geometry and context alignment is learned implicitly—no explicit BEV calibration is required.

\subsection{Channel-Adaptive Semantic Module (CASM)}
A core challenge in cooperative fusion over unreliable links is \emph{semantic aliasing}: when the external stream is degraded, noise makes representations from distinct concepts less separable. CASM addresses this via two steps: (i) it encodes instantaneous link indicators (SNR and modulation) into a multi-token \emph{Channel Prompt} $\mathbf{P}_{\mathrm{ch}}$ using a joint embedding (MLP for continuous SNR, embedding table for discrete modulation, merged by a compact Transformer); and (ii) performing feature-wise adaptive gating. A global gate $\mathbf{g}$ (avg-pool over prompt tokens followed by MLP+Sigmoid) rescales the fused sequence:

\begin{equation}
	\label{eq:casm_icc}
	\mathbf{H}_{\mathrm{enc}}
	= \mathrm{LN}\!\Big(\,\mathbf{S}^{\mathrm{multi}} \odot \big(\mathbf{1}+\mathbf{g}(\mathbf{P}_{\mathrm{ch}})\big)\Big),
\end{equation}
where $\odot$ denotes element-wise multiplication. High SNR yields permissive gates (retain radar-rich semantics); low SNR suppresses unreliable channels and defaults to vision-dominant cues.

\subsection{LLM-Heterogeneous MoE Core \& Decoupled Decoder}
A token-level top-$k$ router dispatches $\mathbf{H}_{\mathrm{enc}}$ to \emph{heterogeneous} MoE~\cite{ref18}---lightweight MLP experts for localization/reconstruction and a temporal Transformer expert for trajectory-with a small load-balancing loss to avoid collapse. The expert outputs, concatenated with $\mathbf{P}_{\mathrm{ch}}$, are fed to a LoRA-tuned GPT-2 that arbitrates task cues and channel context into a globally coherent representation $\mathbf{h}_{\mathrm{LLM}}$. A ViT-U style branch reconstructs images, while a self-attention pooling and two MLP heads regress location and future trajectories. Decoupling mitigates dense–sparse gradient interference and stabilizes joint training.

\subsection{Training Objective}
We train end-to-end with a weighted multi-task loss,
\begin{equation}
	\label{eq:loss_total_icc}
	\mathcal{L}_{\mathrm{total}}
	= \lambda_{\mathrm{rec}}\mathcal{L}_{\mathrm{rec}}
	+ \lambda_{\mathrm{loc}}\mathcal{L}_{\mathrm{loc}}
	+ \lambda_{\mathrm{traj}}\mathcal{L}_{\mathrm{traj}}
	+ \lambda_{\mathrm{vel}}\mathcal{L}_{\mathrm{vel}}
	+ \lambda_{\mathrm{bal}}\mathcal{L}_{\mathrm{bal}},
\end{equation}
where $\mathcal{L}_{\mathrm{rec}}$ combines perceptual terms (e.g., $\ell_1$, SSIM, LPIPS), $\mathcal{L}_{\mathrm{loc}}$ and $\mathcal{L}_{\mathrm{traj}}$ are Smooth-$\ell_1$, $\mathcal{L}_{\mathrm{vel}}$ regularizes first-order differences, and $\mathcal{L}_{\mathrm{bal}}$ promotes uniform expert utilization.

\section{Experimental Evaluation}
We evaluate \emph{LM\textnormal{-}SCIP} on VIRAT and conduct a controlled fusion-gain study on nuScenes. The empirical evaluation comprises three parts: (A) internal validation through ablations; (B) characterization of channel adaptivity; and (C) external positioning via comparisons with state-of-the-art methods.

\begin{table*}[t]
	\caption{Ablation Across Scenarios for LM\textnormal{-}SCIP and Its Variants}
	\label{tab1}
	\centering
	\setlength{\tabcolsep}{2.8pt}      
	\renewcommand{\arraystretch}{1.12} 
	\resizebox{\textwidth}{!}{%
		\begin{tabular}{l ccc ccc ccc ccc}
			\toprule
			\textbf{Scenario} &
			\multicolumn{3}{c}{\textbf{LM-SCIP}} &
			\multicolumn{3}{c}{\textbf{Vision-only}} &
			\multicolumn{3}{c}{\textbf{w/o CASM}} &
			\multicolumn{3}{c}{\textbf{w/o H-MoE}} \\
			\cmidrule(lr){2-4}\cmidrule(lr){5-7}\cmidrule(lr){8-10}\cmidrule(lr){11-13}
			& \textbf{RMSE}$\downarrow$ & \textbf{ADE}$\downarrow$ & \textbf{PSNR}$\uparrow$
			& \textbf{RMSE}$\downarrow$ & \textbf{ADE}$\downarrow$ & \textbf{PSNR}$\uparrow$
			& \textbf{RMSE}$\downarrow$ & \textbf{ADE}$\downarrow$ & \textbf{PSNR}$\uparrow$
			& \textbf{RMSE}$\downarrow$ & \textbf{ADE}$\downarrow$ & \textbf{PSNR}$\uparrow$ \\
			\midrule
			Ideal            & \textbf{0.2140} (\textbf{$\downarrow$42.3\%}) & \textbf{0.1704} & \textbf{22.1755}
			& 0.3708 & 0.2329 & 21.4898
			& 0.2345 & 0.1911 & 22.0684
			& 13.0825 & 1.2398 & 16.0962 \\
			Low SNR          & 0.2435 & 0.1957 & 22.1509
			& 0.3709 & 0.2332 & 21.4866
			& 0.2741 & 0.2240 & 22.0833
			& 77.6543 & 1.2409 & 16.0930 \\
			Occlusion Med.   & 0.3451 & 0.2486 & 19.1614
			& 0.5885 & 0.3770 & 19.1618
			& 0.3536 & 0.2550 & 19.1644
			& 10.2365 & 1.2724 & 15.7033 \\
			Rainy Night      & 0.4390 & 0.2646 & 19.4864
			& 0.5648 & 0.3296 & 19.1367
			& 0.4601 & 0.2844 & 19.3162
			& 46.6108 & 1.2747 & 15.6050 \\
			\bottomrule
	\end{tabular}}
	\vspace{1mm}
	{\footnotesize\emph{Note.} Lower is better for RMSE/ADE ($\downarrow$); higher is better for PSNR ($\uparrow$).}
\end{table*}
\subsection{Experimental Setup}
\textbf{Dataset curation.}
On VIRAT, we adopt an automated pipeline: YOLOv10 for detection/tracking and SAM for instance segmentation. Masked patches are used as visual inputs; raw crops are reconstruction targets. For each instance, a complex I/Q radar waveform is synthesized from its physical state (range,azimuth,velocity) with world coordinates obtained via official homographies.
On nuScenes, we keep the entire recipe fixed and toggle radar availability to isolate fusion gains.

\noindent\textbf{Scenarios and protocol.}
We evaluate four representative conditions on VIRAT: \emph{Ideal} (clean imagery; SNR$=$25\,dB), \emph{Occlusion (medium)} (10--25\% random erasing of the visual patch), \emph{Rainy Night} (image blurring), and \emph{Low SNR + Occlusion} (5\,dB SNR plus medium occlusion). For channel adaptivity, we sweep SNR from $-5$ to $25$\,dB.

\noindent\textbf{Metrics and baselines.}
We report localization (RMSE, meters), trajectory (\(\mathrm{minFDE}_1\)/ADE, meters), and reconstruction (PSNR, dB). Baselines include \textbf{Vision-only}, \textbf{w/o CASM} (no channel awareness), and \textbf{w/o H-MoE} (no task specialization). For operational usability, we also report per-task \emph{QoS accuracy}: fraction of test samples meeting fixed thresholds $\tau_{\text{loc}}{=}0.5$\,m, $\tau_{\text{ADE}}{=}0.5$\,m, and $\tau_{\text{PSNR}}{=}20$\,dB.

\subsection{Ablation Study of Core Components}
\textbf{Table~\ref{tab1}} reports results across four scenarios; ADE and PSNR follow trends consistent with RMSE and are discussed below. These results support three main observations.
\textit{(i) Multimodal fusion is essential.} compared to the Vision-only baseline, the full model reduces RMSE by $42.3\%$ under the \emph{Ideal} setting (from $0.3708$ to $0.2140$\,m).
\textit{(ii) Channel awareness is critical.} Under \emph{Low SNR}, removing \emph{CASM} increases RMSE 
by $12.6\%$ (from $0.2435$ to $0.2741$\,m) and ADE by $14.5\%$ (from $0.1957$ to $0.2240$\,m), highlighting the benefit of explicit channel conditioning. 
\textit{(iii) Task specialization is necessary.} Eliminating the heterogeneous MoE leads to severe degradation; in \emph{Ideal}, RMSE rises from $0.2140$ to $13.0825$\,m, indicating that a monolithic block cannot simultaneously support dense reconstruction and sparse forecasting.
A vision-masked (radar-only) variant further underscores the complementarity of the modalities: in the \emph{Ideal} case, localization errors increase to the metre scale (e.g., RMSE $>2.6$\,m), and image reconstruction becomes infeasible.

In addition to RMSE, we further analyse ADE and PSNR, as well as QoS-oriented accuracies defined at task-specific thresholds.Under the \emph{Ideal} scenario, the full model achieves decimetre-level forecasting (ADE $=0.170$\,m) while maintaining high-fidelity reconstruction (PSNR $>22$\,dB). Under \emph{Occlusion} and \emph{Rainy Night}, ADE increases moderately (to approximately $0.25$--$0.26$\,m) and PSNR decreases by about $3$\,dB, yet radar-aided geometry keeps the predicted motion stable. QoS-oriented accuracy also remains high. In \emph{Ideal}, Loc@0.5\,m, Traj@0.5\,m, and Recon@20\,dB reach $95.97\%$, $93.20\%$, and $85.30\%$, respectively; in \emph{Low SNR}, they remain at $93.77\%$, $90.57\%$, and $85.33\%$. Even under \emph{Occlusion (medium)} and \emph{Rainy Night}, localisation and trajectory accuracies stay in the high--$80\%$ range---$89.17\%/87.62\%$ for Occlusion and $87.75\%/85.51\%$ for Rainy Night---while reconstruction success remains between $53\%$ and $63\%$.

\subsection{Channel-Adaptive Multimodal Fusion Performance}

\begin{figure}[htbp]
	\centering
	\includegraphics[width=0.8\columnwidth]{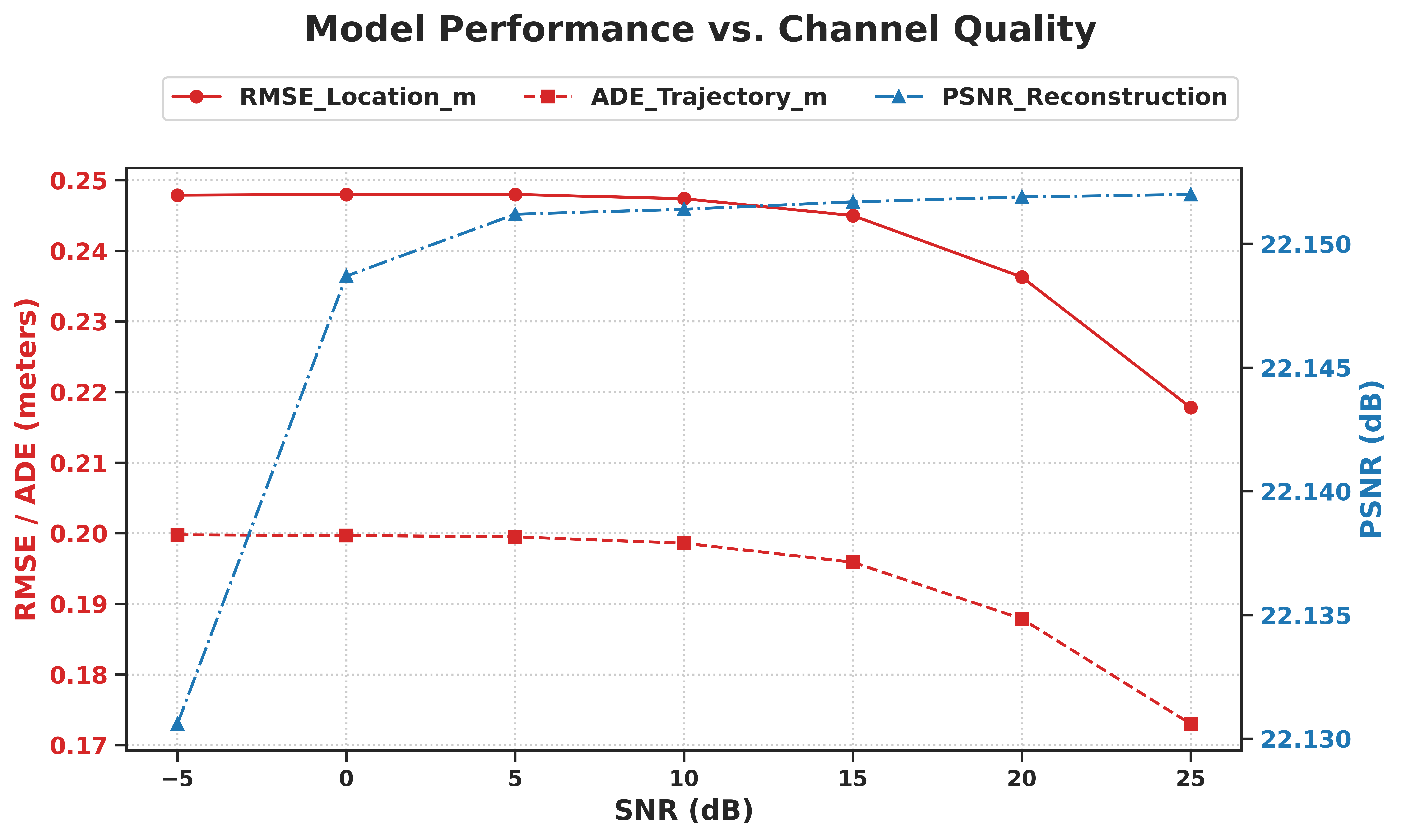}
	\caption{Channel-adaptive performance of \emph{LM\textnormal{-}SCIP} across SNR.}
	\label{fig4} 
\end{figure}

Figure~\ref{fig4} illustrates the performance as a function of SNR in the range from $-5$ to $25$\,dB and exhibits two distinct operating regimes: (\emph{i}) a stable vision-dominant mode at low SNR, where CASM down-weights unreliable radar features; and (\emph{ii}) a synergistic fusion mode at higher SNR, where errors drop markedly (RMSE decreases from $0.245$ to $0.214$\,m and ADE from $0.196$ to $0.170$\,m as SNR increases from $15$ to $25$\,dB). PSNR increases monotonically with SNR, indicating progressively cleaner semantic transmission. These trends are consistent with the learned CASM gating, which suppresses less reliable features at low SNR and becomes more permissive as SNR grows.

\begin{table}[htbp]
	\caption{SOTA comparison. VIRAT: $\mathrm{minFDE}_1$ (m; lower is better). nuScenes: relative $\Delta\mathrm{RMSE}_{\mathrm{Loc}}$ vs.\ vision-only (lower is better).}
	\begin{center}
		\setlength{\tabcolsep}{4pt}
		\renewcommand{\arraystretch}{1.10}
		\begin{tabular}{|l|l|c|}
			\hline
			\multicolumn{3}{|c|}{\textbf{VIRAT} ($\mathrm{minFDE}_1$)} \\
			\hline
			\textbf{Method} & \textbf{Input} & \textbf{Error (m)} \\
			\hline
			Multiverse~\cite{ref19} & Camera-only   & \textbf{0.152} \\
			GraphST~\cite{ref20}    & Camera-only   & 0.153 \\
			SimAug~\cite{ref21}     & Sim-Seg only  & 0.179 \\
			LM\textnormal{-}SCIP (Vision)     & Camera-only     & 0.2299\\
			\textbf{LM-SCIP (Ours)} & \textbf{Camera+Radar} & \textbf{0.179} \\
			\hline
			\multicolumn{3}{|c|}{\textbf{nuScenes} ($\Delta\mathrm{RMSE}_{\mathrm{Loc}}$ vs.\ vision-only)} \\
			\hline
			CenterFusion~\cite{ref22} & Camera+Radar & -9.4\% \\
			CRAFT~\cite{ref3}        & Camera+Radar & -32.3\% \\
			\textbf{LM-SCIP (Ours)}   & \textbf{Camera+Radar} & \textbf{-40.0\%} \\
			\hline
		\end{tabular}
		\label{tab2}
	\end{center}
\end{table}

\subsection{Comparison with State-of-the-Art Methods}

\textbf{Table~\ref{tab2}} summarizes the comparative results. 
First, on VIRAT, \emph{LM\textnormal{-}SCIP} attains \(\mathrm{minFDE}_1 = 0.179\)\,m, matching SimAug (0.179\,m) and close to the best reported camera-only forecasters, while simultaneously delivering high-precision localisation (RMSE $=0.214$\,m), a metric that vision-only trajectory models typically do not report. 
Then, on nuScenes, under a controlled setting with identical training recipes and a radar on/off toggle, adding radar to our framework yields \(\Delta\mathrm{RMSE}_{\text{Loc}}=-40.0\%\) (mATE-equivalent), exceeding the relative gains reported by representative fusion frameworks such as CRAFT.

Taken together, the ablation studies, channel-sweep experiments, and comparisons with state-of-the-art methods show that \emph{LM\textnormal{-}SCIP} (i) provides clear fusion gains under clean conditions, (ii) maintains robust performance under adverse links and visual degradation through CASM-driven re-weighting, and (iii) achieves competitive or superior accuracy compared with leading forecasting models, while uniquely offering high-precision localization and high-fidelity reconstruction within a single channel-aware framework.

\section{Conclusion}
We presented \emph{LM\mbox{-}SCIP}, a channel-aware, LLM-centric multimodal framework that lifts fusion to the semantic layer for infrastructure-centric cooperative perception. By coupling a hierarchical radar--vision encoder with CASM---which embeds V2X link indicators as a ``Channel Prompt''---and a LoRA-tuned LLM with a heterogeneous MoE (H\mbox{-}MoE), the system adaptively balances external radar against local vision and remains stable across sensing/link variations. Experiments on nuScenes and VIRAT show consistent fusion gains, while ablations verify the essential roles of CASM (channel-conditioned re-weighting) and H\mbox{-}MoE (robust reasoning). \emph{LM\mbox{-}SCIP} offers a deployment-friendly path to cooperative, link-aware perception. Future work includes real-world V2X prototypes, richer link side information in CASM, and multi-agent collaboration via prompt-level sharing under dynamic links.

\bibliographystyle{IEEEtran}  
\bibliography{refs} 
\end{document}